\documentclass[runningheads]{llncs}
\usepackage[T1]{fontenc}
\usepackage{xcolor}
\usepackage{chngcntr}
\counterwithout{figure}{section}
\counterwithout{table}{section}
\usepackage{pgfplotstable}
\pgfplotsset{compat=1.18} 
\usepackage{booktabs}
\usepackage{cite}
\usepackage{graphicx}
\usepackage{multirow}
\usepackage{amsmath, amssymb}
\usepackage{bbold}
\usepackage{epstopdf}
\usepackage{subfiles}
\usepackage{lipsum}
\usepackage{tikz}
\usepackage{graphicx}

\usetikzlibrary{calc,bayesnet,arrows}

\providecommand{\xvec}{\ensuremath{\mathbf{x}} }

\providecommand{\Xmat}{\ensuremath{\mathbf{X}} }
\providecommand{\yvec}{\ensuremath{\mathbf{y}} }

\providecommand{\Ymat}{\ensuremath{\mathbf{Y}} }

\DeclareMathOperator*{\argmax}{arg\,max}

\usepackage{url}
\usepackage{subfiles}

\begin{document}

\title{Single-Example Learning in a Mixture of GPDMs with Latent Geometries}

\author{ Jesse St. Amand\inst{1} \and Leonardo Gizzi\inst{2} \and  Martin A. Giese\inst{1}}
\authorrunning{J. St. Amand et al.}

\institute{ Section Computational Sensomotorics,\\  Department of Cognitive Neurology,\\ T\"{u}bingen University Hospital \and  Fraunhofer IPA, \\Department of Biomechatronic Systems} 


\maketitle
\begin{abstract}
We present the Gaussian process dynamical mixture model (GPDMM) and show its utility in single-example learning of human motion data. The Gaussian process dynamical model (GPDM) is a form of the Gaussian process latent variable model (GPLVM), but optimized with a hidden Markov model dynamical prior. The GPDMM combines multiple GPDMs in a probabilistic mixture-of-experts framework, utilizing embedded geometric features to allow for diverse sequences to be encoded in a single latent space, enabling the categorization and generation of each sequence class. GPDMs and our mixture model are particularly advantageous in addressing the challenges of modeling human movement in scenarios where data is limited and model interpretability is vital, such as in patient-specific medical applications like prosthesis control. We score the GPDMM on classification accuracy and generative ability in single-example learning, showcase model variations, and benchmark it against LSTMs, VAEs, and transformers. \vspace{-5pt}
\vspace{-10pt}
\end{abstract}
\section{Introduction}\vspace{-5pt}

Modeling human motion remains a significant challenge across neuroscience, medicine, computer graphics, and robotics due to data scarcity, high dimensionality, motor redundancy, and the requirement for real-time computation \cite{WolpertGhahramani2000}. While neural networks are commonly employed, their effectiveness depends on large training sets that are often difficult or impossible to obtain, particularly in specialized applications. Gaussian processes (GPs), namely the Gaussian process dynamical model (GPDM), offer an effective alternative by enabling accurate human movement synthesis from limited data \cite{Wang2008}.

The GPDM is an extension of the Gaussian process latent variable model (GPLVM) with a hidden Markov model (HMM) prior. Decomposed into two GPs, the emission GP, like in the standard GPLVM, maps the latent space to the observed data, while the dynamical GP models the temporal evolution of the latent space as an HMM. It is effective in low-dimensional representation, motion modeling, and sequence prediction \cite{Wang2008}. Additionally, sparse approximation techniques have made GPDMs both scalable and suitable for real-time applications \cite{Snelson2005,lawrence07_sparse}. Since GPDMs are non-parametric, they are more robust to model misspecification than parametric models, and can better leverage limited data.

GPDMs also offer superior interpretability compared to black-box neural models. Its latent space provides a visualizable low-dimensional representation, its structure is easy to explain with a straightforward Markovian prior, and its probabilistic formulation explicitly quantifies uncertainty in both the dynamics and observations \cite{Wang2008}. These features are critical for applications requiring transparency such as medical diagnosis or prosthesis control.

Despite the many advantages of GPDMs, they have distinct weaknesses. They cannot explicitly classify movements (though they may do so implicitly by predicting the dynamics of a particular class correctly). Furthermore, our own experiments show that they struggle to robustly handle the prediction of multiple movement types simultaneously. This limitation stems from their small temporal prediction horizon, which causes ambiguity when movements intersect or converge in the latent space, leading to inappropriate morphing or switching between classes. Additionally, we found that even when modeling a single movement type, GPDMs yield unstable predictions over longer time horizons due to their step-by-step Markovian prediction process, which accumulates errors that cause trajectories to drift from their true paths.

To address the challenge of modeling diverse movements with limited training examples, we propose the Gaussian process dynamical mixture model (GPDMM), which applies the mixture-of-experts approach \cite{Jacobs1991} to GPDMs. The GPDMM comprises a probabilistic mixture of dynamical GPs, where each GP acts as an expert on a particular movement class's dynamics, all unified by a single emission GP for positional representation. This architecture enables classification while preventing unintended switching or morphing between movements from these different classes. Central to the model's effectiveness is our method for embedding geometric features in the latent space, which enables smooth dynamics and stable long-horizon prediction. Significantly, we achieve high-quality performance using training data with only a single example per movement class, fulfilling one of our primary objectives.

In this paper, we present the formulation of the GPDMM and demonstrate its ability to classify and generate movements. We showcase the model's performance under design variations and ablations, and benchmark it against transformers, VAEs, and LSTMs.
\vspace{-5pt}
\section{Background}
\subsection{Gaussian Process Dynamical Models}
The GPLVM \cite{Lawrence2005} is a non-linear dimensionality reduction approach that maps each high-dimensional data point onto a lower-dimensional latent space via a GP prior. Building on the GPLVM, the GPDM \cite{Wang2008} adds a Markovian prior over the latent states, modeling temporal structure and enabling the generation of new sequences. Furthermore, it allows us to evaluate how likely a new sequence is to have arisen from the trained distribution:
\begin{align}
\label{eq:mmXstarPrior_a}
p\bigl(\Xmat^* \mid a, \Xmat_{in}, \Xmat_{out}\bigr)
\;\propto\;
\exp\!\Bigl(-\tfrac12\,\mathrm{tr}\bigl[\mathbf{K}_{\Xmat^*}^{(a)-1}\,\mathbf{Z}_\Xmat^{(a)}\,\mathbf{Z}_\Xmat^{(a)T}\bigr]\Bigr),
\end{align}
where $\Xmat^*$, $\Xmat_{in}$ and $\Xmat_{out}$ represent, respectively, the latent projection of the new data, and the latent input and output of the autoregressive dynamics. $\mathbf{K_{\Xmat^*}}$ is a kernel matrix evaluated between $\Xmat_{in}$ and $\Xmat^{*}$. $\mathbf{Z}_\Xmat$ is a mean function that includes $\Xmat_{in}$, $\Xmat_{out}$, and $\Xmat^{*}$. Here, $a$ is used as an indicator variable to specify a single GPDM in a mixture model. See Wang et al. \cite{Wang2008} for a full evaluation of the GPDM and eq. \eqref{eq:mmXstarPrior_a}.

The computational cost of full GP inference scales cubically, $\mathcal{O}(N^3)$, where $N$ is the number of training points. To mitigate this, sparse Gaussian process approximations introduce a smaller set of $M$ inducing points ($M \ll N$) \cite{Snelson2005}. The fully independent training conditional (FITC) approach approximates the GP prior so that function values become conditionally independent given these inducing points, reducing computational cost to $\mathcal{O}(NM^2)$ while preserving much of the GP's flexibility \cite{lawrence07_sparse}. Although the data sizes used in these experiments are small enough to employ the full GPDMM with little cost, we build the FITC approximated GPDMM to measure its level of underfitting compared to the full GPDMM and to guide future research on GPDMMs with larger datasets.
\subsection{Mixture of Experts}

Multiple GPDMs can be integrated in a single model using a mixture-of-experts formulation \cite{Jacobs1991}, wherein each dynamical GP is trained only on a subset of the data determined by its class, $S_a \subset S$ where $a \in \{1,2,...,A\}$, $A$ is the number classes, $|S_a| = n_a$, $|S| = N$, and $N$ is the total number of training data points. For our applications, $n_1 = n_2 = ... = n_A = \frac{N}{A}$. Compared to a singular GPDM, this reduces the computational complexity of the dynamical component from $\mathcal{O}(N^3)$ to $\mathcal{O}(\frac{N^3}{A^2})$.

Given a subset of consecutive data points in a sequence (e.g. the first few seconds of a movement), the model can assign likelihoods for each possible category. In the following, we abbreviate eq. \eqref{eq:mmXstarPrior_a} by $p(\Xmat^*|a)$. We use Bayes theorem to evaluate the posterior probabilities,
\begin{equation}
\label{eq:mmBayes}
p(a | \Xmat^*) = \frac{p(a)p(\Xmat^*|a)}{\sum\limits_{\alpha \ \epsilon \ \mathcal{A}} p(\alpha)p(\Xmat^*|\alpha)}, \qquad p(a)=\frac{n_a}{N}
\end{equation} 
where $\mathcal{A}=\{1,2,...,A\}$. A prediction is made by evaluating the argument of the maximum,
\begin{equation}
\label{eq:mmMax}
\argmax_{\alpha \ \epsilon \ \mathcal{A}}{p(\alpha | \Xmat^*)}
\end{equation}
\vspace{-15pt}

\vspace{-10pt}
\section{Methods}
\subsection{Model Architecture}
The basic GPDMM, like GPDMs, employs a single-layer architecture. The observation space of the model represents the observed variables (here joint-angle time series) and is denoted by a single data structure $\Ymat = [\yvec_1, \ldots, \yvec_N]^T \in \mathbb{R}^{N\times D}$ where $D$ is the number of features and $N$ is the number of data points over all sequences. Each column in $\Ymat$ is composed of individual sequences stacked end-to-end for the length of the full data set.

The core of the model is an emission GP that maps a low-dimensional latent space to a higher-dimensional observable data space. Dynamics in the latent space are modeled as state transitions by $G$ dynamical GPs, one for each category, g, of actions. Both the emission and dynamical GPs can be sparsified for scaling to large data sets. Furthermore, the model can be expanded to multiple layers of GPLVMs, though it proved inefficient for our limited data sets (see table \ref{tab:model-variations}).
\vspace{-15pt}
\subsubsection{Implementation.}
The GPDMM was built on top of the GPy library \cite{gpy2014}. In particular, we used their core GPLVMs, kernel functions, and optimization procedures. The mixture model and the bulk of the dynamical GPs were produced in-house with some code heavily adapted from GPy, e.g., in applying sparse approximations. The full (non-sparse) model was about 7.5K parameters on the BM data set and 7.2K on the CMU data set.
\label{fig:model}
\begin{figure}
	\centering
	\includegraphics[width=7cm, height=5cm]{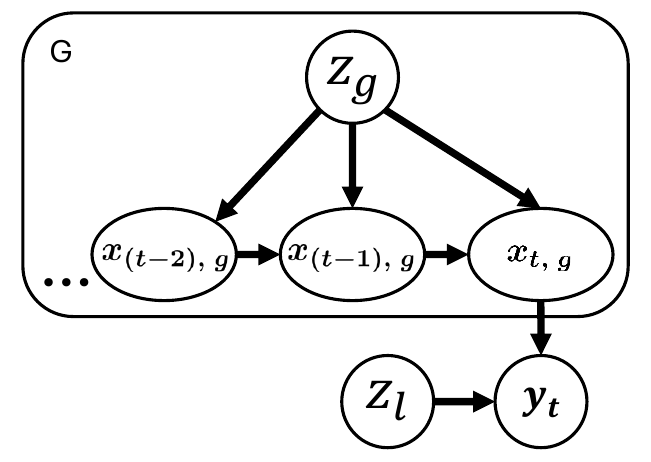}
	\caption{\footnotesize \textit{GPDMM Graphical Model.} 
 The data space represented by a single data structure $\Ymat$ comprises kinematics with $N$ data points consisting  of equal-length sequences with $D$ feature dimensions. The data is represented by a sparse emission GP that maps from a lower-dimensional latent vector $\xvec_t$ at any point in $t$ with inducing inputs $\mathbf{Z_l}$ representing the entire latent space. The latent dynamics are modeled as state transitions by $G$ sparse dynamical GPs with inducing inputs $\mathbf{Z_g}$, each specializing in a distinct action category, $g$. Both $\mathbf{Z}$ variables are dropped for the non-sparse, full-inference model. \vspace{-15pt}}
 
\end{figure}
\vspace{-15pt}
\subsubsection{Comparison to Switching GPDMs.}
The GPDMM shares conceptual ground with Chen et al.'s switching GPDM \cite{Chen2009}. However, it differs on several key points: first, the GPDMM emphasizes stable, long-horizon generation rather than short-term tracking. Second, it uses a fully probabilistic mixture-of-experts structure in a shared latent space, removing the need for discrete switching. Third, our model leverages direct likelihood evaluation for classification and generation, facilitating efficient inference. Crucially, the GPDMM supports single-example learning, embeds geometric features for smoother latent representations, and enables sparse approximations to scale effectively to larger datasets.

\vspace{-10pt}
\subsection{Initial Conditions and Latent Space Geometric Features}\label{sec:latent-geometries}

The optimization of GPLVMs is a highly non-convex problem, making solutions vulnerable to converging on local optima. Consequently, performance can depend significantly on the choice of the initial conditions for the optimization, particularly for the latent space \cite{Bitzer, LiuLosey2017}. Latent space initialization is often done using a form of dimensionality reduction like PCA.

Previous work has employed back-constraints (BCs) to enforce specific topologies in latent spaces. For example, Urtasun et al. \cite{Urtasun2007} used BCs to represent periodic data on a unit circle, and Taubert et al. \cite{taubert2020} extended this idea to non-periodic motions in a hierarchical model. In our single-layer framework for dynamics, we compared both BC-based and unconstrained approaches, ultimately finding better performance with the unconstrained option (see table \ref{tab:model-variations}).
\vspace{-10pt}
\subsubsection{Fourier Basis as Latent Geometries.}
The "unconstrained" approach embeds a geometry in the latent space via initialization with geometric features. Outlined below is our method for constructing these features, using the example of our top-performing geometry based on Fourier basis functions.  

We defined a matrix of latent features $\mathbf{X_G}$ as a set of Fourier basis functions:
\begin{equation}
\mathbf{X_G} = \mathbf{f}(\boldsymbol{\theta}) = 
\begin{bmatrix}
\mathbf{1}, & \cos(2\pi\boldsymbol{\theta}), & \sin(2\pi\boldsymbol{\theta}), & \ldots, & \cos(m\pi\boldsymbol{\theta}), & \sin(m\pi\boldsymbol{\theta})
\end{bmatrix},
\label{eq:fourier}
\end{equation}
where $\mathbf{1}$ is a vector of ones, and $\cos(\cdot)$/$\sin(\cdot)$ terms are included at frequencies from $1$ to $m$. This setup yields a $(2m+1)$-dimensional representation. The hyperparameter $m$ and the inclusion of the constant term were both optimized during model selection.
\vspace{-10pt}
\subsubsection{Mapping Data to the Latent Space.}
We construct a one-dimensional progression vector $\boldsymbol{\theta}$ for each sequence, running from $0$ to $2\pi$; the step size between consecutive data points is inversely proportional to the velocity of trajectory at this point. Thus, points with higher velocities in the original data receive smaller $\boldsymbol{\theta}$ increments, improving coverage in regions sampled sparsely over time. By applying eq. \ref{eq:fourier} to these cumulative $\boldsymbol{\theta}$ values, we obtain a set of $d$-dimensional basis vectors that form the matrix $\mathbf{X_G}$. 
\vspace{-10pt}
\subsubsection{Initialization of Latent Variables.}
Finally, we combine $\mathbf{X_G}$ with features from a suitable dimensionality reduction of the data (see table \ref{tab:latent-geometries}), denoted $\mathbf{X_R}$, to initialize the latent variables:
$\mathbf{X} = \bigl[\mathbf{X_G}, \mathbf{X_R}\bigr] \in \mathbb{R}^{N \times Q}$
where $N$ is the total number of data points and $Q$ is the combined dimension of $\mathbf{X_G}$ and $\mathbf{X_R}$. This initialization preserves the geometric relationships derived from the Fourier basis while leveraging low-dimensional embeddings of the original dataset.
\vspace{-10pt}
\subsubsection{Model Influence.} These geometric embeddings significantly boost performance (see table \ref{tab:latent-geometries}). We reason that this is explained by enforcing smoothness, capturing key periodicities, and allocating finer resolution to high-velocity segments, enabling coherent, comparable representations for diverse motion tempos within a shared latent space. 
\vspace{-10pt}
\subsection{Benchmarking Model Architectures}
The authors are unaware of a standard benchmark for the task of solving classification and long-horizon sequence prediction on small data sets of human movement. Given these limitations, we benchmarked the GPDMM against three popular sequence-generation and classification models: a VAE, Transformer, and LSTM network. Hyperparameter optimization, including variations of the model architectures, was performed consistently for all models (see section \ref{sec:sims}). 
\vspace{-10pt}
\subsubsection{Variational Autoencoder (VAE).}
Implemented in PyTorch \cite{kingma2013auto,paszke2019pytorch}, the VAE features an encoder--decoder design with a reparameterizer (for mean and log variance) plus a classifier branch. It optimizes a combined loss of mean-squared error (reconstruction), KL divergence (latent space regularization), and cross-entropy (classification). This architecture was optimized over the number of hidden and latent dimensions. The number of parameters of the optimized VAE was roughly 12.3M and 7.6M, respectively, for the BM and CMU data sets.
\vspace{-10pt}
\subsubsection{Transformer.}
Built via Hugging Face's BERT architecture \cite{devlin2018bert,wolf2020transformers}, the Transformer applies separate linear heads for classification (using the [CLS] token) and sequence generation. It uses cross-entropy for classification and mean-squared error for generation, optimized with Adam. This architecture was optimized over the number of hidden dimensions, layers, attention heads, and dropout rate. The number of parameters of the optimized Transformer was 19M and 15M, respectively, for the BM and CMU data sets.
\vspace{-10pt}
\subsubsection{Long Short-Term Memory (LSTM).}
An LSTM layer (via PyTorch \cite{hochreiter1997long,paszke2019pytorch}) feeds two linear heads: one for classification (final hidden state) and one for generation (all hidden states). Again, training combines cross-entropy and mean-squared error losses. This architecture was optimized over the number of layers and hidden dimensions. The number of parameters of the optimized LSTM was about 220K and 330K, respectively, for the BM and CMU data sets.
\vspace{-10pt}

\subsection{Resources}
\vspace{-2pt}
The implementation of our proposed methods is publicly available at:\\ \url{https://github.com/jesse-st-amand/H-GPDM-and-MM-GPy-Ext}.

All computations were performed on a single workstation with a AMD Ryzen 7 3700X 8-Core Processor and an NVIDIA GeForce RTX 2070 SUPER graphics card.

Claude 3.7 Sonnet \cite{claude3_7} and ChatGPT o1 \cite{chatgpt_o1} were used in revising the text and for assistance in coding. The authors take full responsibility for the content of the manuscript.
\vspace{-5pt}
\section{Results}
\subsection{Data}
We used two motion-capture data sets: (1) the \emph{Carnegie Mellon University (CMU) data set} \cite{CMUMocap}, comprising 6 trials each of 8 full-body movements (e.g., kicking, lunging, swimming), and (2) our \emph{bimanual (BM) data set}, recorded in-house, featuring upper-body activities of daily living (e.g., lifting a box, rotating a tablet, opening a jar), performed while seated, and aligned to a universal starting position. Each sequence was converted to joint-angle representation at equal-length intervals. The CMU data contained 77 joint features (neck, spine, pelvis, arms, legs); the BM data contained 117 (upper body plus finger articulation).

\vspace{-5pt}
\subsection{Hyperparameter Optimization and Significance Testing}\label{sec:sims}
Bayesian hyperparameter optimization was performed individually per data set via the Python package Scikit-Optimize on all models used in our experiments. Within each iteration of the Bayesian search, we performed a limited Monte Carlo cross-validation (MCCV). At each iteration of the MCCV, we trained on a single example per class and validated and tested on the remaining sequences. For the BM set, this yielded 4 validation and 5 test examples per class; for the CMU set, 2 validation and 3 test examples per class. The validation sets were used to find the optimal number of training epochs before overfitting occurred. We averaged these optimal validation scores over MCCV iterations as input to our Bayesian model evaluation. Final model evaluation and significance testing was performed on the test sets.

Classification used only the first $T$ elements of each test sequence, and generation was performed on the remaining elements. $T$ was set to 40\% and 15\% of the sequence length for the BM and CMU data sets, respectively.
\vspace{-5pt}
\subsection{Scoring Procedure}
To evaluate the GPDMM against other approaches, we computed metrics for both classification accuracy and sequence generation quality.

For scoring classification accuracy, we used the standard F1 score, calculating precision and recall across all classes. 

To measure distances from the ground truth, we computed the Fréchet distance, which measures the distance between trajectories by minimizing the maximum point-wise distance between re-parameterizations of the curves, providing a natural control for time warping \cite{DriemelHarPeledWenk2010}. To account for differing classification accuracy, we only measured sequences that were correctly classified by that model.

We averaged and normalized our distance measures according to the following equation:
\begin{equation}
D_{avg} = \frac{1}{|C|} \sum_{c \in C} \frac{1}{|S_{c,v}|} \sum_{\{\mathbf{s}_{t,a}, \mathbf{s}_{t,b}\} \in S_{c,v}} D_N\bigl(g(\mathbf{s}_{t,a}), \mathbf{s}_{t,b}\bigr)
\label{eq:dist-score}
\end{equation}
\begin{equation}
D_N(\mathbf{s}_g, \mathbf{s}_t) = \frac{d_F(\mathbf{s}_g, \mathbf{s}_t)}{\max_{\mathbf{s}_1, \mathbf{s}_2 \in \mathbf{S}_c} d_F(\mathbf{s}_1, \mathbf{s}_2)}
\label{eq:dist-frac}
\end{equation}
where $C$ is the set of all classes, $S_{c}$ is the set of all testing set sequences in class $c$, $S_{c,v}$ is the subset of those sequences correctly classified, $d_F(\mathbf{s}_1, \mathbf{s}_2)$ is the Fréchet distance between $\mathbf{s}_1$ and $\mathbf{s}_2$, $D_N(\cdot)$ is the normalized Fréchet distance, $\mathbf{s}_{t,a}$ is the ground truth sub-sequence used in classification, $\mathbf{s}_{t,b}$ is the remaining ground truth sub-sequence, and $\mathbf{s}_{g}$ is the sequence generated by $g(\mathbf{s}_{t,a}) = \mathbf{s}_{g}$ of size $\mathbf{s}_{t,b}$. This normalization controls for biases between sequence classes.

In addition to distance, we evaluated dampening and smoothness metrics. Dampening captures the reduction in motion amplitude and follow-through compared to natural movements. Smoothness quantifies the fluidity of a movement trajectory, with poor scores indicating jittery movements. These metrics address quality dimensions that F1 and distance metrics can miss. See table \ref{tab:model-variations} row 7 (Fourier; None) for an example of a model that scored well on all metrics except for dampening, and table \ref{tab:model-comparison} row 3 (VAE) for a similar example that scores poorly on LDJ. 

For the dampening metric, we analyze mean displacement over sliding windows calculated as,
\begin{equation}\label{eq:dampening}
d = \frac{1}{N-w} \sum_{i=1}^{N-w} \left\| \mathbf{p}_{i+w} - \mathbf{p}_i \right\|
\end{equation}
where $N$ is the number of points in the sequence, $w$ is the window size, and $\mathbf{p}_i$ is the position of the $i$-th point. The dampening metric is computed as a ratio between ground truth and generated sequences. Dampening scores greater than 1 indicate that generated movements exhibit diminished amplitude or incomplete execution compared to the ground truth. 

To quantify smoothness, we employed the log dimensionless jerk (LDJ) metric \cite{Balasubramanian2011}, which controls for duration and amplitude. The LDJ is defined as:
\begin{equation}\label{eq:ldj}
\eta_{LDJ} = -\ln\left(\frac{(t_2-t_1)^3}{\nu_{\text{peak}}^2} \cdot \int_{t_1}^{t_2} \left(\frac{d^3x}{dt^3}\right)^2 dt\right)
\end{equation}
where $t_2-t_1$ represents the movement duration, $\nu_{\text{peak}}$ is the peak velocity, and the integral represents the squared jerk over the movement interval. Higher (less negative) LDJ values indicate smoother movements. Similar to the dampening metric, we calculate the LDJ ratio between generated and ground truth sequences, where values greater than 1 indicate degraded smoothness in the generated movements.


\vspace{-5pt}
\subsection{Experiments} \label{sec:experiments}

Tables~\ref{tab:latent-geometries}, \ref{tab:model-variations}, and \ref{tab:model-comparison} summarize model performances with \textit{F1} indicating the F1 score, and \textit{Distance}, \textit{Dampening}, and \textit{LDJ} represented by eqs. \eqref{eq:dist-score}, \eqref{eq:dampening}, and \eqref{eq:ldj}, respectively. An asterisk~(*) denotes a statistically significant difference from the top performing model for that metric ($p < 0.05$). Values in bold indicate the best score per column. Values not significantly different from the top performing model are considered competitive. For both the dampening and LDJ metrics, we consider scores closer to 1 to be better since this indicates a value closer to the ground truth. However, in-context we found LDJ scores below 1 to often be acceptable since this indicates trajectories with "less noise" than the ground truth--a possible indicator for over-smoothing that is (in)validated by the dampening metric. \vspace{-10pt}

\begin{table}[ht]
\caption{GPDMM comparison for top-performing variations of geometries (geo) and dimensionally reduced features (DR) as initial conditions in the latent space. Dims refers to the number of latent dimensions occupied by the geometry where "NA" is not applicable and "all" is the total number of dimensions in the latent space.}
\label{tab:latent-geometries}
\centering
\small
\resizebox{\textwidth}{!}{%
\begin{tabular}{|l|l|cc|cc|cc|cc|cc|}
\hline
 &  & \multicolumn{2}{c|}{\textbf{Dims}} & \multicolumn{2}{c|}{\textbf{F1}} & \multicolumn{2}{c|}{\textbf{Distance}} & \multicolumn{2}{c|}{\textbf{Dampening}} & \multicolumn{2}{c|}{\textbf{LDJ}} \\
\textbf{Geo} & \textbf{DR} & \textbf{BM} & \textbf{CMU} & \textbf{BM} & \textbf{CMU} & \textbf{BM} & \textbf{CMU} & \textbf{BM} & \textbf{CMU} & \textbf{BM} & \textbf{CMU} \\
\hline
None & Isomap & NA & NA & 0.781$^{*}$ & 0.996 & 0.233$^{*}$ & 0.135$^{*}$ & 3.551 & 1.398 & 1.026 & 0.902$^{*}$ \\
None & kPCA & NA & NA & \textbf{0.959} & \textbf{1.0} & 0.232$^{*}$ & 0.181$^{*}$ & 6.44$^{*}$ & 43.719 & 0.784$^{*}$ & 0.754$^{*}$ \\
None & PCA & NA & NA & 0.941 & \textbf{1.0} & 0.175$^{*}$ & 0.148$^{*}$ & 1.925$^{*}$ & 3.551$^{*}$ & 0.882$^{*}$ & 0.785$^{*}$ \\
None & Random & NA & NA & 0.829$^{*}$ & 0.769$^{*}$ & 0.359$^{*}$ & 0.338$^{*}$ & 94.513$^{*}$ & 37.257$^{*}$ & \textbf{0.998} & \textbf{0.992} \\
None & UMAP & NA & NA & 0.909 & 0.996 & 0.172$^{*}$ & 0.138$^{*}$ & 3.523$^{*}$ & 1.716$^{*}$ & 0.881$^{*}$ & 0.772$^{*}$ \\
Chebyshev & None & All & All & 0.943 & \textbf{1.0} & 0.145 & 0.126$^{*}$ & 1.625$^{*}$ & 1.782$^{*}$ & 0.813$^{*}$ & 0.816$^{*}$ \\
Fourier & None & All & All & 0.938 & \textbf{1.0} & \textbf{0.131} & 0.11 & 1.624$^{*}$ & 1.297$^{*}$ & 0.807$^{*}$ & 0.77$^{*}$ \\
Laguerre & None & All & All & 0.911 & \textbf{1.0} & 0.152 & 0.136$^{*}$ & 1.639$^{*}$ & 1.915$^{*}$ & 0.73$^{*}$ & 0.784$^{*}$ \\
Ellipse & kPCA & 2 & 2 & 0.955 & \textbf{1.0} & 0.151$^{*}$ & 0.116 & 1.526$^{*}$ & 1.084$^{*}$ & 0.775$^{*}$ & 0.718$^{*}$ \\
Fourier & kPCA & 11 & 3 & 0.941 & \textbf{1.0} & 0.152$^{*}$ & 0.115$^{*}$ & \textbf{1.104} & 1.048$^{*}$ & 0.943$^{*}$ & 0.788$^{*}$ \\
Fourier & PCA & 13 & 4 & 0.928$^{*}$ & \textbf{1.0} & 0.174$^{*}$ & \textbf{0.107} & 1.214$^{*}$ & 1.046 & 1.005 & 0.768$^{*}$ \\
Laguerre & Isomap & 3 & 2 & 0.806$^{*}$ & \textbf{1.0} & 0.236$^{*}$ & 0.135$^{*}$ & 2.24 & 1.179$^{*}$ & 1.013 & 0.884$^{*}$ \\
Torus & PCA & 3 & 3 & 0.946 & \textbf{1.0} & 0.167$^{*}$ & \textbf{0.107} & 1.582$^{*}$ & 1.062$^{*}$ & 0.859$^{*}$ & 0.764$^{*}$ \\
Torus & UMAP & 3 & 3 & 0.92 & 0.965$^{*}$ & 0.145$^{*}$ & 0.124$^{*}$ & 1.495$^{*}$ & \textbf{1.002} & 0.868$^{*}$ & 0.8$^{*}$ \\
\hline
\end{tabular}
}\vspace{-1pt}
\end{table}

\begin{table}[ht]
\caption{GPDMM parameter comparison for top-performing variations of the numbers of layers, order of the dynamics, and back-constraints (BCs).}
\label{tab:model-variations}
\centering
\small
\begin{tabular}{|l|l|l|cc|cc|cc|cc|}
\hline
 &  &  & \multicolumn{2}{c|}{\textbf{F1}} & \multicolumn{2}{c|}{\textbf{Distance}} & \multicolumn{2}{c|}{\textbf{Dampening}} & \multicolumn{2}{c|}{\textbf{LDJ}} \\
\textbf{Layers} & \textbf{Order} & \textbf{BC} & \textbf{BM} & \textbf{CMU} & \textbf{BM} & \textbf{CMU} & \textbf{BM} & \textbf{CMU} & \textbf{BM} & \textbf{CMU} \\
\hline
1 & 1 & None & 0.941 & \textbf{1.0} & \textbf{0.152} & \textbf{0.106} & \textbf{1.104} & 1.041 & 0.943 & 0.769 \\
1 & 2 & None & 0.918 & \textbf{1.0} & 0.169 & 0.108 & 1.108 & 1.055 & 0.936 & 0.788 \\
2 & 1 & None & 0.945 & \textbf{1.0} & 0.17 & 0.107 & 1.331$^{*}$ & \textbf{1.026} & 0.914 & 0.857 \\
1 & 1 & Kernel & 0.924 & \textbf{1.0} & 0.167 & 0.108 & 1.169 & 1.073 & 0.928 & 0.78 \\
1 & 1 & GP & 0.935 & \textbf{1.0} & 0.171 & 0.108 & 1.369 & 1.066 & \textbf{0.949} & 0.784 \\
1 & 1 & Linear & \textbf{0.963} & \textbf{1.0} & 0.171 & 0.145$^{*}$ & 1.88$^{*}$ & 2.213$^{*}$ & 0.77$^{*}$ & 0.782 \\
1 & 1 & MLP & 0.934 & \textbf{1.0} & 0.212$^{*}$ & 0.176$^{*}$ & 1.752 & 11.658 & 0.881 & 0.766 \\
1 & 1 & C-Kernel & 0.348$^{*}$ & 0.261$^{*}$ & 0.448$^{*}$ & 0.331$^{*}$ & 1.883$^{*}$ & 3.776 & 0.992 & \textbf{0.884} \\
\hline
\end{tabular}
\vspace{-12pt}
\end{table}

\begin{table}[ht]
\caption{Comparison between the GDMM and benchmark models.}
\label{tab:model-comparison}
\centering
\small
\begin{tabular}{|l|cc|cc|cc|cc|}
\hline
 & \multicolumn{2}{c|}{\textbf{F1}} & \multicolumn{2}{c|}{\textbf{Distance}} & \multicolumn{2}{c|}{\textbf{Dampening}} & \multicolumn{2}{c|}{\textbf{LDJ}} \\
\textbf{Model} & \textbf{BM} & \textbf{CMU} & \textbf{BM} & \textbf{CMU} & \textbf{BM} & \textbf{CMU} & \textbf{BM} & \textbf{CMU} \\
\hline
GPDMM & 0.93 & \textbf{1.0} & \textbf{0.155} & \textbf{0.108} & \textbf{1.101} & \textbf{1.032} & \textbf{0.945} & 0.77$^{*}$ \\
Sparse GPDMM & \textbf{0.936} & \textbf{1.0} & 0.191$^{*}$ & 0.113$^{*}$ & 1.882$^{*}$ & 1.145$^{*}$ & 0.807$^{*}$ & 0.726$^{*}$ \\
VAE & 0.874$^{*}$ & \textbf{1.0} & 0.189$^{*}$ & 0.131$^{*}$ & 1.328$^{*}$ & 1.037 & 1.262$^{*}$ & 1.228$^{*}$ \\
Transformer & 0.903$^{*}$ & 0.944$^{*}$ & 0.356$^{*}$ & 0.254$^{*}$ & 3.318$^{*}$ & 4.642$^{*}$ & 0.854$^{*}$ & 0.761$^{*}$ \\
LSTM & 0.534$^{*}$ & 0.871$^{*}$ & 0.292$^{*}$ & 0.206$^{*}$ & 345.211 & 548.459$^{*}$ & 0.818$^{*}$ & \textbf{0.886} \\
\hline
\end{tabular}
\vspace{-12pt}
\end{table}
Table \ref{tab:latent-geometries} illustrates how adding geometric and dimensionally reduced (DR) features to the latent space initialization impacts GPDMM performance. The table shows only top-performing configurations from a much larger set of parameters tested, including additional geometries and DR methods. All geometries were constructed as described in section \ref{sec:latent-geometries}. The ellipse and torus were created using their parametric equations. The Chebyshev and Laguerre geometries were defined analogously to the example Fourier geometry in eq. \eqref{eq:fourier}. The column labeled "dims" indicates how many latent dimensions a given geometry occupied. Function-based geometries (e.g., Fourier), which lack a fixed dimensionality, were optimized as hyperparameters. From these experiments, we concluded that combining a Fourier geometry with a form of PCA produced the best set of results, particularly due to the consistently strong dampening scores on both data sets compared to other approaches.

Table \ref{tab:latent-geometries} also emphasizes the importance of the dampening metric. Many variations of the GPDMM and our benchmarks generated movements that scored well on distance and LDJ, but when evaluated by eye, were severely under-expressing their range of motion or halting in place. Dampening scores above around 1.3 were found to be indicative of poor performance. Furthermore, in row 4 (None; Random), the GPDMM scores well on LDJ, but very poorly on dampening, indicating an artificially good LDJ score (further examination revealed the generated motion to be both under-expressed and noisy).

Table \ref{tab:model-variations} reports the GPDMM's performance under variations in the layer count, order of the dynamical model, and type of back-constraint (BCs). The first row presents the GPDMM's performances with the Fourier-PCA latent spaces from table \ref{tab:latent-geometries}. The subsequent rows detail our most successful models optimized over hyperparameters (e.g. latent dimensions, geometries, and DRs) for specified numbers of layers, orders, and BCs. The terms "GP," "MLP," and "C-Kernel" respectively refer to Gaussian process, multi-layer perceptron, and circular kernel BCs. Based on these findings, we concluded that a simplified single-layer, first-order-dynamics model without BCs performed best.

Table \ref{tab:model-comparison} compares the GPDMM and its sparse variant ($M = N/2$) against our three benchmark models: the VAE, LSTM, and transformer. On the BM data set, the GPDMM maintained strong performance across all metrics, significantly outperforming the other approaches in distance, LDJ, and dampening, and performing within significance of the highest F1 score produced by the sparse model. Performances on the CMU data set closely matched the BM data set. Again, the GPDMM performed well on all metrics, scoring significantly above the other models on distance. It tied with the sparse model, the transformer, and the VAE for the best F1 score, and scored within significance of the VAE on dampening. While the LSTM achieved the top LDJ value, its extremely high dampening makes this score unreliable.

\begin{figure}\label{fig:latent_spaces}
	\centering
	\includegraphics[width=12cm, height=6cm]{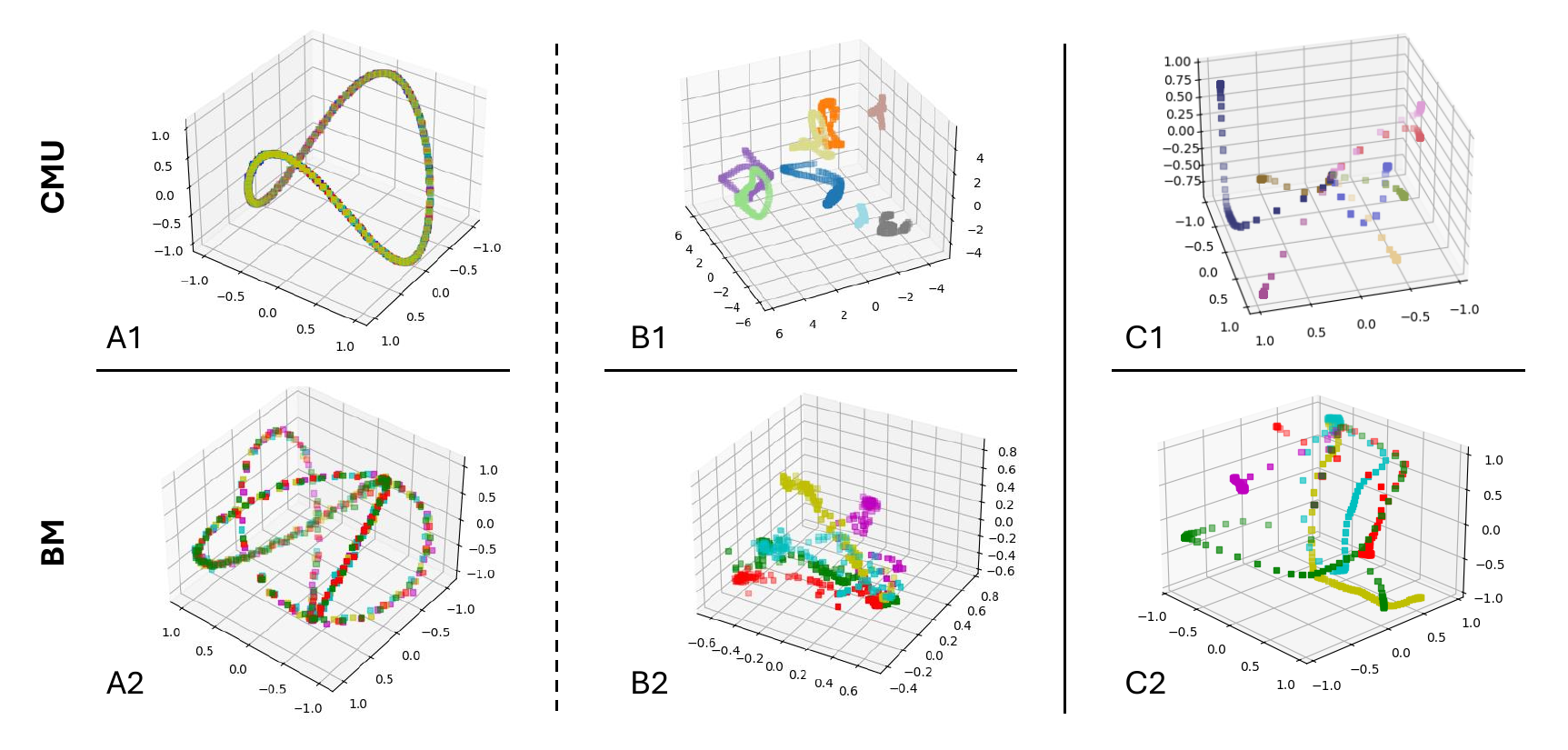}
	\caption{\footnotesize \textit{GPDMM and LSTM Latent Space Visualizations.} The figure displays latent spaces for select feature dimensions of top-performing GPDMMs and LSTMs. The top plots (1) display models trained on the CMU data set, while the bottom plots (2) show the BM data set. The left-hand and middle plots (A and B) depict GPDMM latent spaces. "A" plots illustrate the Fourier basis geometries. "B" plots present latent features that were initialized by dimensionally reduced representations of the data, "B1" with standard PCA and "B2" with RBF kernel PCA. The right-hand plots (C) show the first three dimensions of the LSTM's hidden state dynamics. \vspace{-15pt}}
\end{figure}
Figure 2 presents feature dimensions of the GPDMM's latent space and the LSTM's hidden state dynamics when separately trained on the CMU (the top figures marked with 1) and BM (bottom, marked 2) data sets. Plots A show the GPDMM's geometric representations for capturing dynamics. A1 shows the first three Fourier basis functions (eq. \ref{eq:fourier}), and A2 shows functions 4-6. Different function sets were selected for each plot as the models produce visually similar latent representations for identical functions. Data points are aligned with the geometry and distributed sequentially across its breadth to promote accurate state transitions (see section \ref{sec:latent-geometries}). Plots B show features optimized for class discrimination and exhibit more variation across the space. Plots C display the first three dimensions of the LSTM's hidden state dynamics, combining information on dynamics and class discrimination together. B2, C2, and C1 display a divergence from a common centerpoint. This divergent pattern accurately represents the movement structure in the BM dataset but not in the CMU dataset. The GPDMM approach correctly captures this distinction (B1), while the LSTM misrepresents it (C1).

\vspace{-5pt}
\section{Discussion}
We introduced the GPDMM, a mixture-of-experts framework that leverages geometry-embedded latent spaces to classify and generate diverse motion classes from minimal data. Across two human-motion datasets, our model consistently outperformed or matched popular neural baselines, while preserving the interpretability inherent to GPDMs. These results highlight the ability of the GPDMM to produce robust, data-efficient motion analysis and generation in scenarios such as prosthetic control, where patient-specific data sets are  limited in size, and reliability and transparency are critical.
\begin{credits}
\subsubsection{\ackname} The authors thank the International Max Planck Research
School for Intelligent Systems (IMPRS-IS) for supporting Jesse St. Amand. This research was funded through the European Research Council ERC 2019-SYG under EU Horizon 2020 research and innovation programme (grant agreement No. 856495, RELEVANCE). The CMU mocap data set used in this project was obtained from mocap.cs.cmu.edu and was created with funding from NSF EIA-0196217.
\vspace{-7pt}
\subsubsection{\discintname}
The authors have no competing interests to declare that are
relevant to the content of this article.
\end{credits}

\bibliographystyle{splncs04}
\bibliography{main}
\end{document}